\documentclass[runningheads]{llncs}
\usepackage{graphicx}
\usepackage{
cite}
\usepackage{hyperref}
\usepackage{amsmath,amssymb,amsfonts}
\usepackage{algorithmic}
\usepackage{graphicx}
\usepackage{textcomp}
\usepackage{multirow}
\usepackage{amsmath}
\usepackage[misc]{ifsym}
\usepackage{amssymb}
\usepackage{xcolor}

\newcommand{\bc}{\mathbf{c}}\newcommand{\bC}{\mathbf{C}}
\newcommand{\bD}{\mathbf{D}}

\newcommand{\bI}{\mathbf{I}}
\newcommand{\bJ}{\mathbf{J}}
\newcommand{\bK}{\mathbf{K}}

\newcommand{\bM}{\mathbf{M}}

\newcommand{\bO}{\mathbf{O}}
\newcommand{\bP}{\mathbf{P}}
\newcommand{\bq}{\mathbf{q}}
\newcommand{\bR}{\mathbf{R}}
\newcommand{\bS}{\mathbf{S}}
\newcommand{\bt}{\mathbf{t}}\newcommand{\bT}{\mathbf{T}}

\newcommand{\bx}{\mathbf{x}}\newcommand{\bX}{\mathbf{X}}

\newcommand{\bmu}{\boldsymbol{\mu}}

\newcommand{\bSigma}{\boldsymbol{\Sigma}}

\DeclareMathOperator*{\argmin}{argmin~}

\makeatletter
\DeclareRobustCommand\onedot{\futurelet\@let@token\@onedot}
\def\@onedot{\ifx\@let@token.\else.\null\fi\xspace}

\makeatother

\newcommand{\jx}[1]{\textcolor{black}{#1}}

\begin{document}
\title{Free-SurGS: SfM-Free 3D Gaussian Splatting for Surgical Scene Reconstruction}

\titlerunning{SfM-free 3D Gaussian Splatting for Surgical Scene Reconstruction}
\author{Jiaxin Guo\inst{1} \and
Jiangliu Wang\inst{1}\and
Di Kang\inst{2} \and
Wenzhen Dong\inst{1} \and \\ Wenting Wang\inst{1} \and Yun-hui Liu\inst{1,3}\textsuperscript{(\Letter)}}
\authorrunning{J. Guo et al.}
\institute{The Chinese University of Hong Kong, Hong Kong SAR, China  \and
Tencent AI Lab, Shenzhen, China \and
Hong Kong Center for Logistics Robotics, Hong Kong SAR, China
}
\maketitle              %
\begin{abstract}
Reconstructing surgical scenes plays a vital role in computer-assisted surgery,
holding a promise to enhance surgeons' visibility.
Recent advancements in 3D Gaussian Splatting (3DGS) have shown great potential for real-time novel view synthesis of general scenes, which relies on accurate poses and point clouds generated by Structure-from-Motion (SfM) for initialization.
However, 3DGS with SfM fails to recover accurate camera poses and geometry in surgical scenes due to the challenges of minimal textures and photometric inconsistencies.
To tackle this problem, in this paper, we propose the first SfM-free 3DGS-based method for surgical scene reconstruction by jointly optimizing the camera poses and scene representation.
 Based on the video continuity, the key of our method is to exploit the immediate optical flow priors to guide the projection flow derived from 3D Gaussians. 
Unlike most previous methods relying on photometric loss only, we formulate the pose estimation problem as minimizing the flow loss between the projection flow and optical flow. 
A consistency check is further introduced to filter the flow outliers by detecting the rigid and reliable points that satisfy the epipolar geometry.  
During 3DGS optimization, we randomly sample frames to optimize the scene representations to grow the 3D Gaussians progressively.
Experiments on the SCARED dataset demonstrate our superior performance over existing methods in novel view synthesis and pose estimation with high efficiency. Code is available at \url{https://github.com/wrld/Free-SurGS}.

\keywords{Novel View Synthesis \and 3D Reconstruction \and 3D Gaussian Splatting \and Endoscopic Surgery.}
\end{abstract}
\section{Introduction}
Reconstructing surgical scenes is crucial for revealing internal anatomical structures during minimal invasive surgery (MIS), and enables many
downstream applications such as augmented reality, virtual reality, surgical planning, and surgical simulation~\cite{chen2018slam, liu2020reconstructing, tang2018augmented}. 
While neural radiance fields (NeRF)~\cite{mildenhall2021nerf} methods demonstrate success for novel view synthesis from multiple photos or videos, their applicability is limited for computational efficiency in training and inference. 
Recently, 3D Gaussian Splatting (3DGS)~\cite{kerbl20233d}, which introduces anisotropic 3D Gaussians to build explicit scene representations, emerges as a powerful rendering technique for its rendering efficiency and the ability to produce high-fidelity images. 
3DGS showcases significant potential in advancing novel view synthesis, offering a promising pathway to establish real-time, interactive surgical simulations.
\begin{figure*}[t]
\centering
\includegraphics[width=1.0\linewidth]{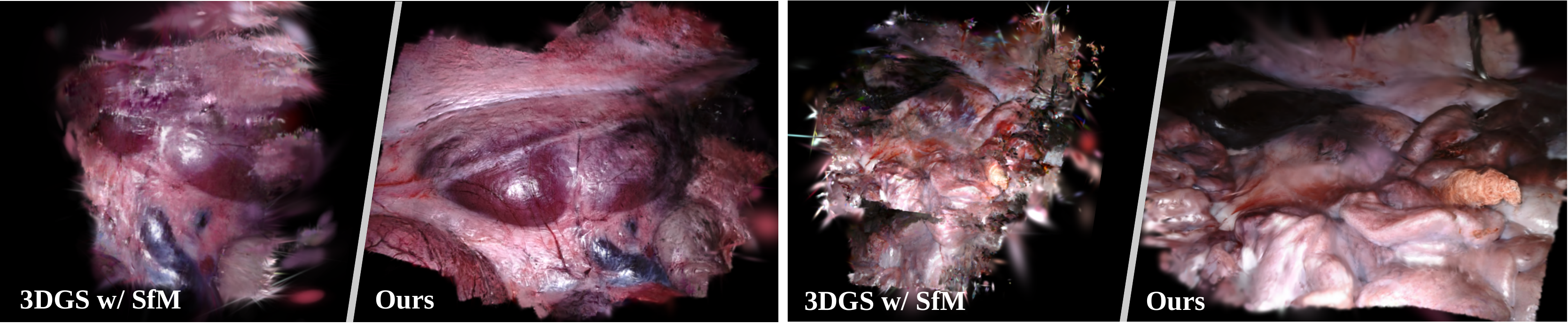}
\caption{
\textbf{3DGS~\cite{kerbl20233d} meets a major limitation in its reliance on SfM.} We propose Free-SurGS to eliminate this need and demonstrate better performance.
}
\label{fig: teaser}
\end{figure*}

Despite the advances, 3DGS encounters a major limitation in its reliance on the camera poses and sparse point clouds from Structure-from-Motion (SfM)~\cite{schonberger2016structure}, which inevitably influences its application in surgical videos. 
This pre-processing stage is too time-consuming to run for long sequence endoscopic videos, limiting their employment in inter-operative applications. 
Furthermore, SfM is prone to fail on the appearance of surgical scenes that contain minimal surface textures and photometric inconsistencies like non-Lambertian surfaces, reflective surfaces, and illumination fluctuation. 
This creates difficulties in detecting features for correspondence search, leading to pose estimation failure and point clouds from incomplete views. 
As shown in Fig. \ref{fig: teaser}, taking the inaccurate poses and point clouds for initialization, the 3D Gaussians show floaters and artifacts in the rendered images and reconstruct incorrect geometry. 
To address this issue, some SfM-free studies~\cite{wang2021nerf,lin2021barf,bian2023nope, fu2023colmap, jeong2021self} are proposed to reduce or eliminate the reliance on SfM by estimating the camera poses along with optimizing the scene representations. However, most approaches optimize the camera poses by minimizing the photometric loss between the rendered image and input frame, leading to inaccurate pose estimation due to the homogeneity of textures and photometric inconsistencies.

In this paper, we address the challenges and present Free-SurGS for \jx{fast surgical scene reconstruction and real-time rendering from monocular inputs}, realizing joint optimization for both 3D Gaussians and camera poses. 
However, the challenges of the appearance in surgical scenes motivate us to exploit the optical flow priors based on video continuity to guide the projection flow derived from the 3D Gaussians. 
Our contribution is summarized as threefold: 
\textbf{1)} We present the first SfM-free 3DGS-based approach for fast surgical scene reconstruction and real-time rendering from monocular inputs only. 
\textbf{2)} Unlike previous methods relying on photometric loss only, we formulate the pose estimation problem as matching the projection flow derived from 3D Gaussians with optical flow. A consistency check is further proposed to detect the rigid and reliable points that are consistent with the epipolar geometry. 
\textbf{3)} The extensive experimental results on the SCARED datasets demonstrate that our method outperforms the existing methods in both novel view synthesis and pose estimation, achieving photo-realistic surgical scene rendering with real-time inference speed.

\section{Methodology}
\begin{figure*}[t]
\centering
\includegraphics[width=1.0\linewidth]{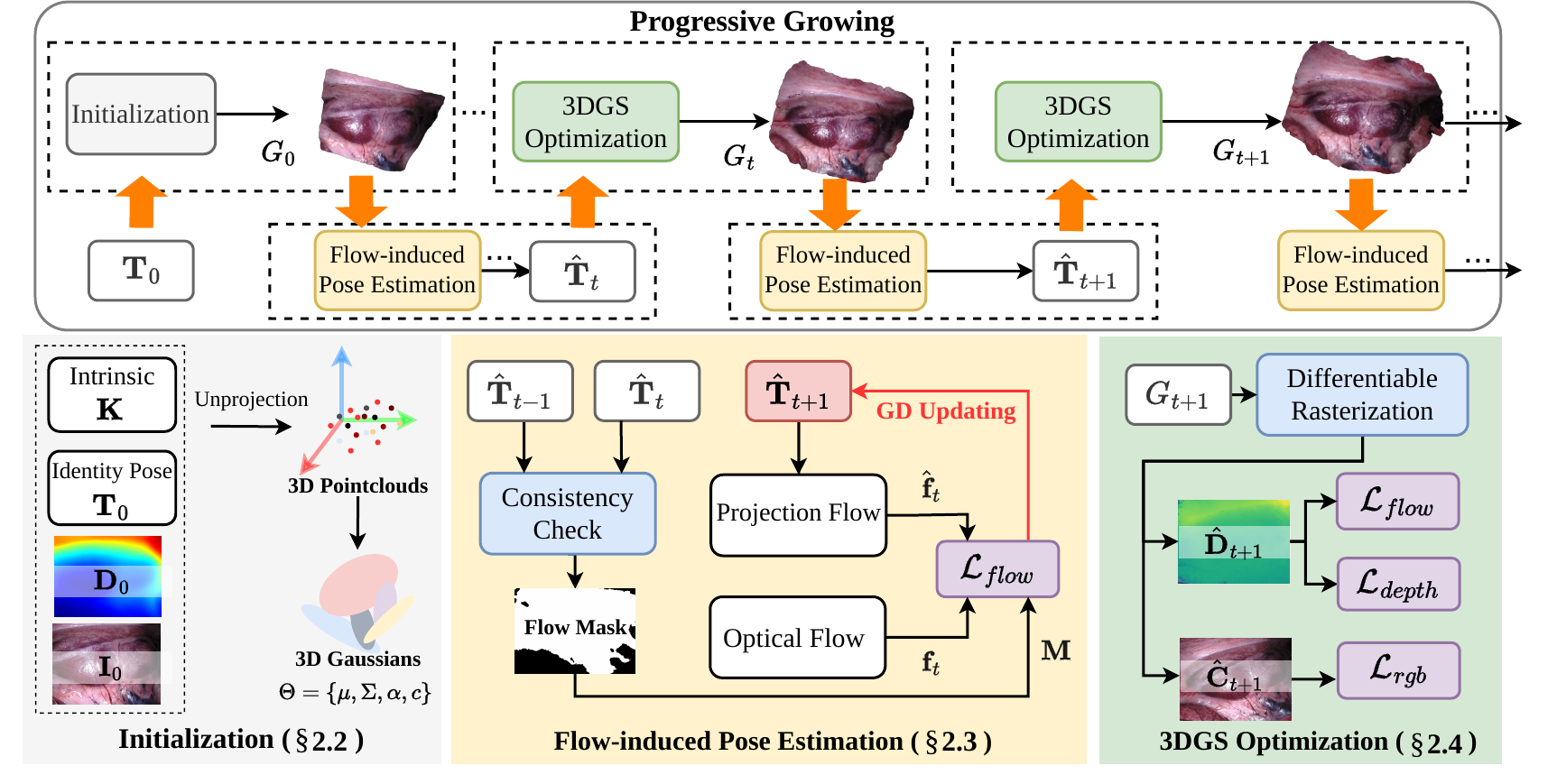}
\caption{\textbf{Overview of our proposed Free-SurGS.} Given endoscopic monocular images as input, we jointly estimate the camera poses and optimize the 3D Gaussians iteratively by progressive growing. }
\label{fig: method}
\end{figure*}
In this paper, we model the surgical scene as 3D Gaussians to render photo-realistic images from free viewpoints. 
Given a sequence of monocular images $\{ \bI_0, \dots, \bI_{N-1}\}$ shot by a moving endoscope,
our goal is to better reconstruct the complete surgical scene via a joint optimization of the camera poses and the 3D representation (i.e. 3DGS).

\jx{Given the input image sequence, we utilize off-the-shelf methods to obtain the monocular depth $\{ \bD_t\}_{t=0}^{N-1}$ from Depth-Anything~\cite{yang2024depth} and optical flow between $\bI_t$ and $\bI_{t+1}$ as $\{ \bO_{t \rightarrow t+1}\}_{t=0}^{N-1}$ from RAFT~\cite{teed2020raft} as pseudo-GT.} As shown in Fig.~\ref{fig: method}, we first initialize the 3D Gaussians $G_0$ from the frame $\bI_0$ utilizing the point clouds from monocular depth $\bD_{0}$ and the identity camera pose $\bT_0$ (Sec. \ref{sec: initial}).
Based on the continuity of surgical video, the 3D Gaussian is updated from every input image consequently following a progressive growing process.
We formulate the pose estimation problem as guiding the projection flow of 3D Gaussians with the robust correspondences from $\bO_{t\rightarrow t+1}$ under a consistency check, to compensate for the limitation of photometric loss (Sec. \ref{sec: pose_estimation}). During 3DGS optimization, we randomly sample frames with estimated poses to optimize the scene representation (Sec. \ref{sec: gaussian_optimize}).

\subsection{Preliminary: 3D Gaussian Splatting}\label{sec: 3dgs}
3DGS~\cite{kerbl20233d} introduces the 3D Gaussians as differential volumetric representations of radiance fields, allowing high-quality real-time novel view synthesis. The set of 3D Gaussians is initialized from the calibrated camera poses and sparse point clouds generated from SfM. Each Gaussian is defined by position $\bmu$, covariance matrix $\bSigma$: $G(\bx) = e^{-\frac{1}{2}(\bx-\bmu)^T\bSigma^{-1}(x-\bmu)}.$
The covariance can be decomposed from a scaling matrix $\bS$ and rotation matrix $\bR$: $\bSigma = \bR \bS \bS^T \bR^T$. To render a novel view, the 3D Gaussians are projected to 2D camera view $\bT$: $\bSigma' = \bJ \bT \bSigma \bT^T \bJ^T$, where $\bJ$ is the Jacobian of the affine approximation of the projective transformation. To render the color, 3DGS further optimizes opacity and SH coefficients, following the point-based differential rendering by rasterizing anisotropic splats with $\alpha$-blending. The color and depth are rasterized following:
\begin{equation}
    \hat{\bC} = \sum_i^N \bold{c}_i \alpha_i \prod_j^{i-1} (1-\alpha_j), \quad \hat{\bD} = \sum_i^N d_i \alpha_i \prod_{j}^{i-1}(1-\alpha_j),
\end{equation}
where $\bold{c}_i$ and $\alpha_i$ denote the color and opacity of the Gaussian, $d_i$ is the z-axis of the points by projecting the center of 3D Gaussians $\bmu$ to the camera coordinate. In summary, the parameters to optimize for the Gaussians include: $\Theta = \{ \bmu, \bSigma, \alpha, \bc \}$. To realize SfM-free scene reconstruction, we need to both recover the camera poses $\bT$ and optimize the Gaussian parameters $\Theta$.

\subsection{Initialization from Monocular Depth}\label{sec: initial}
Given first frame $\bI_0$ and the known intrinsic $\bK$, we generate the pointcloud $\bP$ by unprojecting the monocular depth $\bD_0$ by the initial identity camera pose $\bT_0$: $\bP = \pi^{-1}(\bT_0, \bD_0, \bK)$, where $\pi^{-1}$ is the pixel-to-world projection. The center of Gaussians $\bmu$ is initialized by $\bP$. The color of each point $\bc$ is initialized with the SH coefficient from the first frame. 
Other parameters are initialized following the implementation in 3DGS~\cite{kerbl20233d}. After initialization, we optimize the 3D Gaussians $ G_0$ by minimizing the losses introduced in Sec. \ref{sec: gaussian_optimize}.

\subsection{Flow-induced Pose Estimation}\label{sec: pose_estimation}
In this step, we fix the parameters of 3D Gaussians (i.e. assume the current GS is pseudo-GT) and update the camera pose by matching the projection flow from 3D Gaussians with the robust correspondences from filtered optical flow. 

\noindent\textbf{Pose Estimation via Pointcloud Transformation. }We formulate the camera pose estimation problem into predicting the transformation of 3D Gaussians following \cite{keetha2023splatam, fu2023colmap}. Given the position of Gaussian center $\bmu$, we can project it to 2D camera view $\bT$ by $\bmu_{2D} = \bK\frac{\bT \bmu}{(\bT \bmu)_z}$.
Therefore, the camera pose estimation is equivalent to estimating the transformation of 3D Gaussians.

To update the camera pose by gradient descent, we first transform the 3D Gaussians $G$ with the camera pose $\bT$. We take the camera poses as the optimizable variables and represent the rotations in quaternion $\bq$ and translation vector $\bt$. 
At timestep $t+1$, its camera pose $\hat{\bT}_{t+1}$ is initialized from the previous camera poses $\hat{\bT}_{t}$ and $\hat{\bT}_{t-1}$ based on the constant velocity assumption:
$\hat{\bold{q}}_{t+1} = \hat{\bold{q}}_{t} + (\hat{\bold{q}}_{t}-\hat{\bold{q}}_{t-1}) + \Delta \bold{q}, \quad \hat{\bold{t}}_{t+1} = \hat{\bold{t}}_{t} + (\hat{\bold{t}}_{t}-\hat{\bold{t}}_{t-1}) + \Delta \bold{t}.$

 Previous methods~\cite{bian2023nope, lin2021barf} mostly adopt the photometric loss $\mathcal{L}_{rgb}$ to match the rendered color $\hat{\bold{C}}_{t+1}$ and ground truth color $\bold{I}_{t+1}$ for pose estimation with gradient optimization:
 \begin{equation}
     \mathcal{L}_{rgb} = (1-\lambda)\mathcal{L}_{1} + \lambda \mathcal{L}_{\text{D-SSIM}}.
 \end{equation}
 However, the application of photometric loss in optimizing camera poses within surgical scenes encounters limitations. 
 First, the homogeneity and sparse texturing of the surgical surfaces lead to ambiguities in feature matching. 
 Second, photometric inconsistencies across different views are quite common due to varied lighting conditions, the existence of reflective surfaces of the surgical instruments and tissues, and the presence of non-Lambertian surfaces.
 Consequently, using only the photometric loss for pose estimation is prone to converge to some local minima, thus leading to inaccurate reconstruction in the following step.
 
\begin{figure*}[t]
\centering
\includegraphics[width=1.0\linewidth]{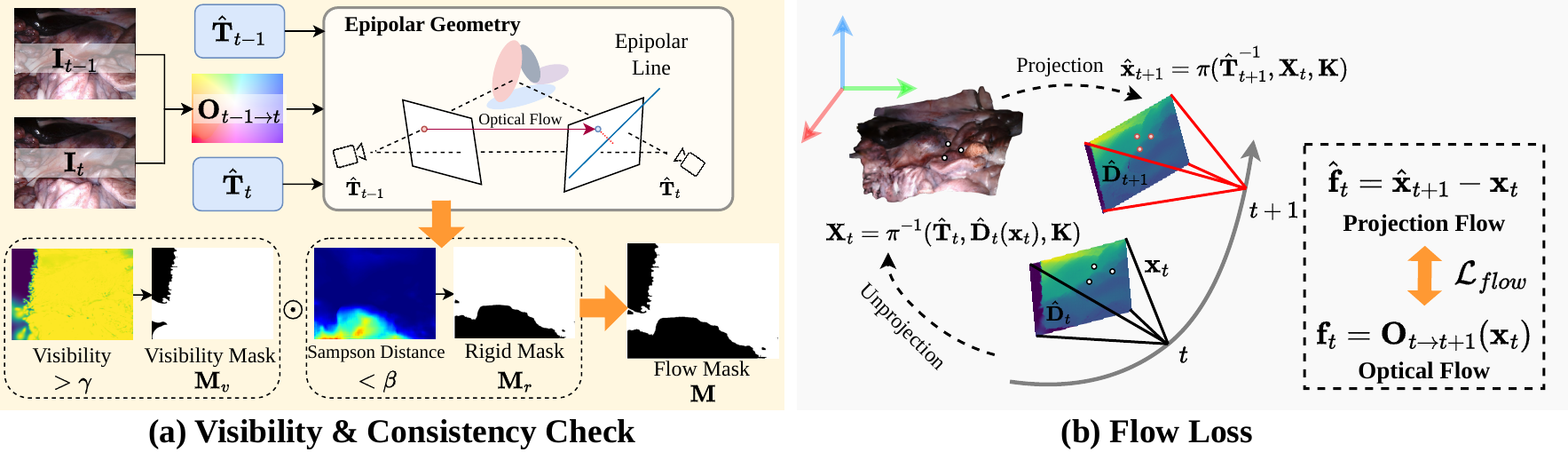}
\caption{
\textbf{Illustration of our proposed flow-induced pose estimation. }
(a) The consistency check is introduced to filter out the outliers in the optical flow map $\bO_{t-1\rightarrow t}$ to obtain reliable and robust correspondences. 
(b) We formulate the pose estimation problem as matching the projection flow with the optical flow, to compensate for the limitations of photometric loss.
}
\label{fig: con_flow}
\end{figure*}

\noindent\textbf{Projection Flow. } 
As shown in Fig. \ref{fig: con_flow}(b), we introduce a projection flow to compute the per-pixel movement by projecting the 3D Gaussians from camera view $\hat{\bT}_{t}$ to $\hat{\bT}_{t+1}$.
Specifically, we first unproject $\bx_t$ (i.e. each pixel of $\bI_t$) to 3D points $\bX_t$ with rendered depth $\bD_t$ and $\hat{\bT}_t$. Next, the correspondences $\hat{\bx}_{t+1}$ can be obtained by projecting $\bX_t$ to camera view $\hat{\bT}_{t+1}$. The projection flow $\hat{\bold{f}}_t$ can be computed by:
\begin{equation}
\begin{aligned}
    \hat{\bold{f}}_t = \hat{\bx}_{t+1} - \bx_t&=\pi(\hat{\bT}_{t+1}^{-1}, \bX_t, \bK) - \bx_t, \\
    \quad \text{where} \quad \bX_t &= \pi^{-1}(\hat{\bT}_{t}, \hat{\bD}_t(\bx_t), \bK).
\end{aligned}    
    \end{equation}
By computing the transformation of 3D Gaussians from one camera view to the next, the projection flow is less dependent on texture variations, making it more reliable in surgical scenes.

\noindent\textbf{{Visibility \&} Consistency Check. }
First, we employ a visibility check to filter the optical flow from the visibility map to exclude not yet constructed regions. 
During the first epoch to learn the scene representation, the 3D Gaussians are partially reconstructed, resulting in empty regions in the rendered view. We compute the visibility map $\bM_{v}$ of 3DGS in the rendered view under $\hat{\bT}_{t+1}$, by accumulating the opacity of Gaussians under camera view $\hat{\bT}_{t+1}$: $\bM_v = \sum_i^N \alpha_i \prod_{j}^{i-1}(1-\alpha_j) > \gamma$, where $\gamma$ is the threshold for visibility. 

Second, a consistency check is introduced to remove the outliers to maintain rigid and reliable points in the optical flow. 
In dynamic surgical environments characterized by transient objects and photometric inconsistencies, it is essential to identify and preserve correspondences that are both rigid and reliable for accurate matching. Utilizing the optical flow $\bO_{t-1 \rightarrow t}$, we assess the epipolar geometry informed by the estimated camera poses $\hat{\bT}_{t-1}$ and $\hat{\bT}_{t}$. This assessment ensures that correctly matched points align with their respective epipolar lines for robust matching.  Therefore, we can find the rigid and reliable points that better satisfy the epipolar geometry in $t$ to further filter out outliers in $\bO_{t\rightarrow t+1}$ based on the continuity of endoscopic video. As shown in Fig. \ref{fig: con_flow}(a), we compute the Sampson distance~\cite{hartley2003multiple} to measure the geometric error between a point in one image and its corresponding epipolar line in the other image. We take a threshold $\beta$ to obtain a rigid mask $\bM_r$ for time $t$, ensuring that only robust correspondences are utilized for subsequent pose estimation tasks from $t$ to $t+1$. 
Finally, we obtain the flow mask from the consistency check: $\bM = \bM_v \odot \bM_r.$

\noindent\textbf{Flow Loss. }To guide the pose estimation from dense correspondence in optical flow $\bO_{t\rightarrow t+1}$, the flow loss is defined by minimizing the $L_2$ loss between the optical flow and projection flow with flow mask $\bM$:
\begin{align}
    \mathcal{L}_{flow} &= \parallel \bM \odot (\hat{\bold{f}_t} - \bold{f}_t)\parallel_2^2, \quad \text{where } \bold{f}_t = \bO_{t\rightarrow t+1}(\bx_t).
\end{align}
The flow loss compensates for the photometric loss to tackle the challenging surgical scene and enhance the pose estimation accuracy:
 \begin{equation}
     \hat{\bT}_{t+1} = \argmin_{\bT_{t+1}} \lambda_1 \mathcal{L}_{rgb} + \lambda_2 \mathcal{L}_{flow},
 \end{equation}
 where $\lambda_1$ and $\lambda_2$ denote the weight for $\mathcal{L}_{rgb}$ and $\mathcal{L}_{flow}$. By addressing both the geometric consistency through $\mathcal{L}_{flow}$ and the photometric similarity through $\mathcal{L}_{rgb}$, our free-GS ensures a more robust alignment of the camera poses, even in the presence of textural homogeneity or photometric anomalies.
\subsection{3D Gaussians Optimization}\label{sec: gaussian_optimize}
After estimating the camera pose $\hat{\bT}_{t+1}$, we optimize the parameters $\Theta$ of 3D Gaussians $G$. Here, we keep the camera pose fixed and optimize the scene representation by minimizing the photometric loss, depth loss, and flow loss:
\begin{equation}
    \hat{\Theta} = \argmin_{\Theta} \lambda_1 \mathcal{L}_{rgb} + \lambda_2 \mathcal{L}_{flow} + \lambda_3 \mathcal{L}_{depth},
\end{equation}
where $\lambda_3$ is the weight for depth loss, $\mathcal{L}_{depth}$ denotes a scale-invariant loss~\cite{Ranftl2022} between rendered depth $\hat{\bD}$ and monocular depth $\bD$ generated from Depth-Anything~\cite{yang2024depth}. Since the projection flow is derived from rendered depth for reprojection, the flow loss directly contributes to a more precise estimation of depth. By optimizing the 3D Gaussians for both photometric consistency and flow dynamics, the geometry of the 3D Gaussians is not only consistent with the observed image data but also adheres to the expected motion patterns across frames. Finally, we add or prune the 3D Gaussians with adaptive density control, resulting in a progressive growing process for reconstruction. 
\section{Experiments}
\subsection{Implementation Details}
\noindent \textbf{Experimental Setup. }All experiments are implemented using Pytorch~\cite{paszke2019pytorch} on NVIDIA RTX 3090 GPU. 
We set the same parameters for all the surgical scenes. The optimizer and hyper-parameters of 3D Gaussians follow the original implementation of 3DGS~\cite{kerbl20233d}. We use Adam optimizer~\cite{kingma2014adam} for pose estimation with a learning rate of $4 \times 10^{-3}$. \jx{During the progressive growing, we set 30 iterations for both pose estimation and 3DGS optimization.}
\begin{figure*}[t]
\centering
\includegraphics[width=1.0\linewidth]{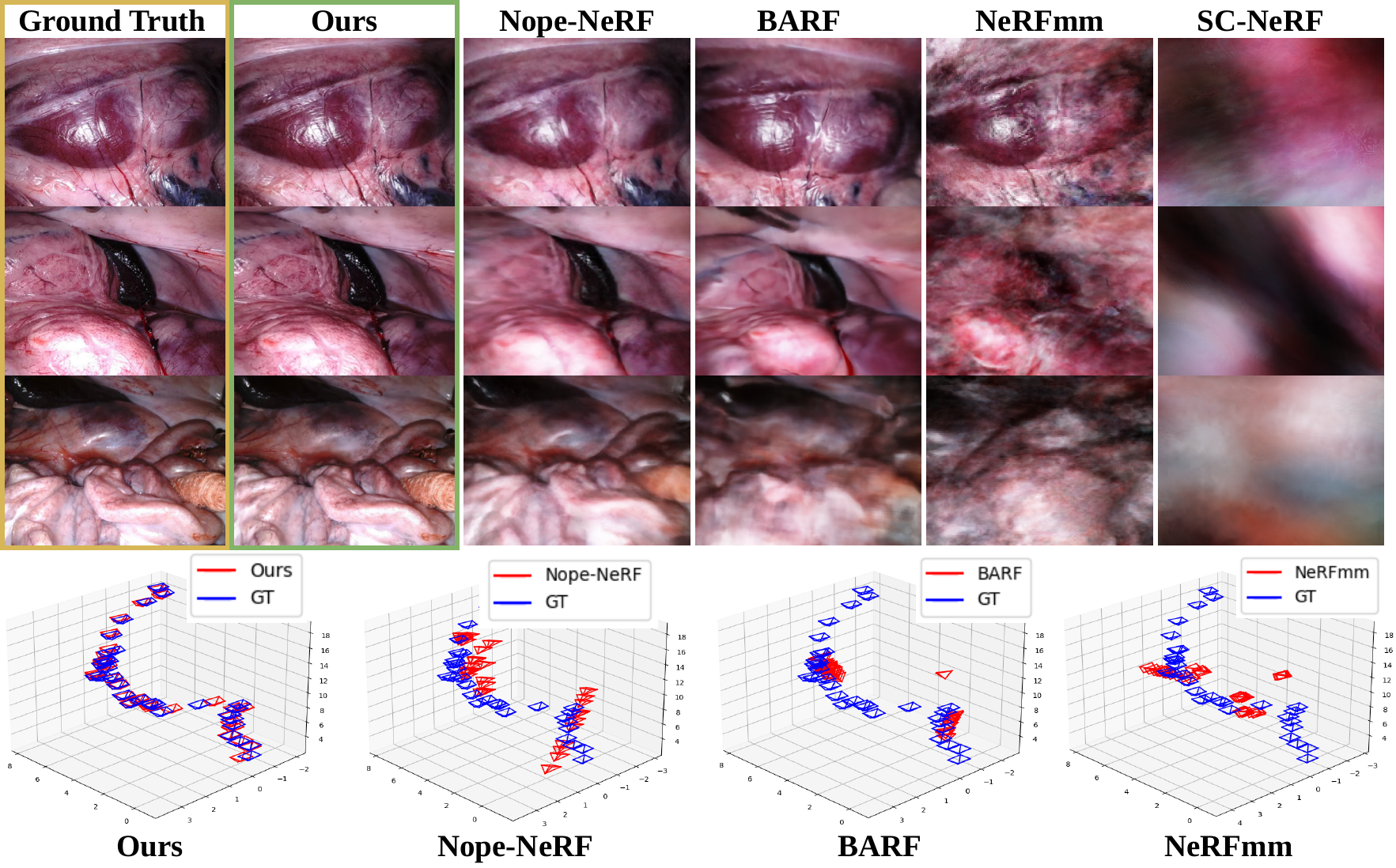}
\caption{\textbf{Qualitative results} of novel view synthesis and pose estimation.}
\label{fig: qual}
\end{figure*}
\begin{table}[tbp]

\centering
\caption{\textbf{Quantitative comparison results }on the SCARED Dataset~\cite{allan2021stereo}.
}
\label{tab: results}
\resizebox{0.95\linewidth}{!}{%
\begin{tabular}{cccccccccccccccc}
\hline
& \multirow{2}{*}{Methods} &  & \multicolumn{3}{c}{Novel View Synthesis} &  & \multicolumn{3}{c}{Pose Estimation} &  & \multicolumn{3}{c}{Efficiency}  \\ 
\cline{4-6} \cline{8-10} \cline{12-14} &  &  & PSNR $\uparrow$  & SSIM $\uparrow$  & LPIPS $\downarrow$   &  & RPE$_t$ $\downarrow$  & RPE$_r$ $\downarrow$  & ATE $\downarrow$ &  & Train $\downarrow$ & FPS $\uparrow$  & GPU $\downarrow$  \\ \hline    
  & SC-NeRF~\cite{jeong2021self}    &  & 9.943 & 0.344 & 0.654 & & 6.436 & 6.802 & 13.67 &  & \underline{5.0 h} & 0.074   & \textbf{3.2 G}    \\
  & NeRFmm~\cite{wang2021nerf}                    &  & 16.55 & 0.361 & 0.540  & &  5.681 & 9.108 & 12.74&  & 9.5 h & 0.27          & 6.0 G    \\
  &BARF~\cite{lin2021barf}                  &  & 16.25 & 0.511& 0.658 & & \underline{5.005} & 6.515         & \underline{9.832} & & 7.2 h &0.12& 8.5 G\\
  &Nope-NeRF~\cite{bian2023nope}                  &  &  \underline{21.42} & \underline{0.620} & \underline{0.523}  & &5.632 & \underline{5.685}   &  12.30  &  & 50.0 h & 0.34  & 8.0 G    \\
    &\textbf{Ours}                  &  & \textbf{24.35} & \textbf{0.741} & \textbf{0.270} & &\textbf{3.299} & \textbf{1.966} & \textbf{5.854}  &  & \textbf{1.0 h} & \textbf{60.0}  & \underline{3.8 G}    \\
\hline
\end{tabular}
}
\end{table}

\noindent \textbf{Datasets.}
We evaluate our approach on the SCARED Dataset~\cite{allan2021stereo}, which is a real-world dataset with challenging endoscopic scenes containing reflective surfaces, illumination fluctuations, and weak textures. The image resolution for training and evaluation is $640 \times 480$ on the SCARED Dataset. We test 9 scenes from the SCARED Dataset with 50-150 frames for each scene with one-eighth of the images for test following \cite{bian2023nope}. The SCARED Dataset also provides ground truth camera poses of every frame for evaluation.
\begin{table}[tbp]

\scriptsize
\centering
\caption{\textbf{Ablation study} of flow-induced pose estimation. \jx{“Con.” refers to the consistency check to maintain rigid and reliable points.}
}
\label{tab: ablation}
\resizebox{0.8\linewidth}{!}{%
\begin{tabular}{ccc|ccccccc}
\hline
$\mathcal{L}_{rgb}$&$\mathcal{L}_{flow}$ & Con.& PSNR $\uparrow$  & SSIM $\uparrow$  & LPIPS $\downarrow$   &  & RPE$_t$ $\downarrow$  & RPE$_r$ $\downarrow$  & ATE $\downarrow$\\ \hline    
$\checkmark$  &  &  & 20.57 & 0.603 & 0.438 & & 8.574 & 4.151 &  10.08\\
  & $\checkmark$ &  & 22.75 & 0.652 & 0.382  & & 4.133 & 2.769 & 7.410    \\
 $\checkmark$ & $\checkmark$&  & 23.53 & 0.688 & 0.291 & & 3.512& 2.438         & 6.435   \\
$\checkmark$  & $\checkmark$& $\checkmark$ & \textbf{24.35} & \textbf{0.741} & \textbf{0.270} & & \textbf{3.299} & \textbf{1.966}      & \textbf{5.854}   \\
\hline
\end{tabular}
}
\end{table}

\noindent \textbf{Evaluation Metrics.}
We evaluate the performance of novel view synthesis via PSNR, SSIM~\cite{wang2004image}, and LPIPS~\cite{zhang2018unreasonable}. To compare the accuracy of estimated camera poses, we evaluate Absolute Trajectory Error (ATE), and Relative Pose Error (RPE), including rotation RPE$_r$ and translation RPE$_t$ following \cite{bian2023nope}. Note that the unit for RPE$_t$ and ATE is millimeter (mm), and the unit for RPE$_r$ is degree.

\subsection{Quantitative and qualitative results}
We compare our method with existing state-of-the-art SfM-free methods: Nope-NeRF~\cite{bian2023nope}, BARF~\cite{lin2021barf}, NeRFmm~\cite{wang2021nerf} and SC-NeRF~\cite{jeong2021self}. Quantitative results in Tab. \ref{tab: results} demonstrate that our method outperforms all the baselines. 
Only based on photometric loss, BARF~\cite{lin2021barf}, NeRFmm~\cite{wang2021nerf}, and SC-NeRF~\cite{jeong2021self} fail to recover the correct camera pose, suffering from the challenging surgical scenes. 
With constraints from depth distortion, Nope-NeRF~\cite{bian2023nope} improves the performance compared to other baselines but still fails to handle large endoscopic movement (See Fig. \ref{fig: qual}). 
Thanks to the flow matching and the consistency check, our Free-SurGS could estimate accurate camera poses for scene reconstruction and render photo-realistic images with 3DGS. 
The efficiency comparison in Tab. \ref{tab: results} also demonstrates our faster training, higher inference speed, and lower memory of parameters, satisfying real-world surgical applications.

We conduct ablation studies to validate the effectiveness of the proposed modules in Tab. \ref{tab: ablation}.
The flow loss $\mathcal{L}_{flow}$ compensates for the limitation of photometric loss and improves the accuracy of pose estimation. 
The consistency check could further enhance the robustness of large movement and semi-static scenes. 
With more accurate poses as input, the performance of 3DGS is further improved to reconstruct the surgical scene.

\section{Conclusion}
In this paper, we propose Free-SurGS as the first SfM-free 3DGS-based method to realize multi-view surgical scene reconstruction. To handle the challenging surgical scene with minimal textures and photometric inconsistencies, we use the optical flow priors to guide the projection flow derived from 3D Gaussians for robust pose estimation. Extensive experiments on the SCARED dataset show that our method outperforms the previous methods in both novel view synthesis and pose estimation, achieving fast reconstruction and real-time rendering with less training time. Our method shows potential to provide a highly realistic and interactive environment that could advance preoperative planning and training practices. However, our method is limited in handling dynamic scenes with severe tissue deformations, which we will address in the future work.

\begin{credits}
\subsubsection{\ackname} This work is supported in part by Shenzhen Portion of Shenzhen-Hong Kong Science and Technology Innovation Cooperation Zone under HZQB-KCZYB-20200089, in part by the Research Grants Council of Hong Kong under Grant T42-409/18-R, Grant 14218322, and Grant 14207320, in part by the Hong Kong Centre for Logistics Robotics, in part by the Multi-Scale Medical Robotics Centre, InnoHK, and in part by the VC Fund 4930745 of the CUHK T Stone Robotics Institute.

\subsubsection{\discintname} The authors have no competing interests to declare that are relevant to the content of this article.
\end{credits}

\bibliographystyle{splncs04}
\bibliography{egbib}

\end{document}